\documentclass{article} % For LaTeX2e
\usepackage[T1]{fontenc}
\usepackage{iclr2027_conference,times}

\usepackage{amsmath,amsfonts,bm}

\def\eqref#1{equation~\ref{#1}}
\def\1{\bm{1}}

\DeclareMathAlphabet{\mathsfit}{\encodingdefault}{\sfdefault}{m}{sl}
\SetMathAlphabet{\mathsfit}{bold}{\encodingdefault}{\sfdefault}{bx}{n}

\usepackage{hyperref}
\hypersetup{pdftitle={Sanity Checks for Attention Surgery: Null Models and Matched Interventions for Representation-Motivated Design},pdfauthor={Basavaraj R M, Aryan Bharat Kumar}}
\usepackage{url}

\usepackage{amsmath}
\usepackage{amssymb}
\usepackage{booktabs}
\usepackage{multirow}
\usepackage{graphicx}

\newcommand{\vhat}{\hat{\mathbf{v}}}
\newcommand{\yv}{\mathbf{y}}
\newcommand{\zv}{\mathbf{z}}

\title{Sanity Checks for Attention Surgery: \\
Null Models and Matched Interventions for \\
Representation-Motivated Design}

\author{\hfil Basavaraj R M\\[3pt]
\hspace{0.3cm}Independent AI \& ML Systems Researcher\\[3pt]
\hspace{0.3cm}\hfil\texttt{basavarajrm972@gmail.com}
\And
\hfil Aryan Bharat Kumar\\[3pt]
\hspace{0.3cm}Independent ML Researcher\\[3pt]
\hspace{0.3cm}\hfil\texttt{ar22073yan@gmail.com}
}

\iclrfinalcopy

\begin{document}

\maketitle
\lhead{Technical report}

\begin{abstract}
Transformer architectural modifications frequently measure a statistic on internal representations, observe that it is large, remove it architecturally, and report an improvement. Interpreting that sequence requires evaluating whether the statistic exceeds its noise floor, whether it survives a geometric null, and whether the training gain requires the specific direction removed. We formulate these requirements as three reusable checks and apply them to self-attention removal, which deletes the component of an attention head's output aligned with its own value vector. Across nine multi-head pretrained models (5,408 heads, 124M to 6.9B parameters), an earlier-position causal null accounts for 48\% to 63\% of the head-value cosine alignment, with no monotone scale trend. Under grouped-query attention, this alignment is specific to each head's own key--value group and reverses across groups (+0.24 to +0.27 within-group, $-0.11$ to $-0.19$ across-group). Split-half reliability demonstrates that the per-head intervention effect is resolvable in GPT-2 ($r_\Delta = 0.80$) but attenuated in Pythia-160M ($0.42$) and Pythia-410M ($0.53$), whereas structural self-cosine reliability is near-perfect in all three ($r_{\text{stat}} > 0.99$); which statistic predicts its own intervention is model-dependent. In a pre-registered 3-arm by 8-seed paired factorial at $4.0\times10^{8}$ tokens per run ($399\,900\,672$ tokens), a fixed arbitrary rank-one direction does not differ from baseline ($+0.0011$ nats, $p = 0.197$), while self-directed removal lowers validation loss ($-0.0029$ nats, Holm-corrected $p = 0.0023$) and outperforms the fixed arbitrary control directly ($-0.0040$ nats, $p = 0.0018$). At 50.9M scale the training benefit requires the own-value direction rather than a fixed arbitrary one, but a large cosine alignment alone does not establish redundancy in frozen models.
\end{abstract}

\section{Introduction}

Multi-head self-attention \citep{vaswani2017attention} forms each head's
output as a weighted sum of value vectors. In a causal model, position $i$
can attend to its own value vector as well as to earlier positions. Exclusive
self attention (XSA) removes the output component aligned with that value
\citep{zhai2026xsa}.

\paragraph{Research objective.}
We examine whether this component is redundant and whether its cosine
alignment identifies heads whose removal changes language-modelling loss.
This is a question about a direction within a head, distinct from the
whole-head pruning studied by \citet{michel2019sixteen,voita2019analyzing}.

Two measurements are needed to interpret such an intervention. First, high
cosine alignment need not be specific to a token's own value:
contextual representations can be anisotropic \citep{ethayarajh2019contextual}.
A comparison with other values tests how much alignment survives that
background geometry. Second, an edit's loss effect must be measured
reliably before it can be related to a structural statistic. A small observed
correlation can reflect either a weak association or noisy measurements.
Training with an arbitrary-direction edit provides a further control for
whether an apparent benefit requires the own-value direction.

We evaluate these questions through three sequential checks:

\paragraph{Check 0: is the effect resolvable?} Before interpreting correlations
between representation statistics and intervention outcomes, we estimate the
split-half reliability $r_\Delta$ of the behavioural effect and $r_{\text{stat}}$
of the structural statistic. Their geometric mean $\sqrt{r_\Delta r_{\text{stat}}}$
provides an attenuation diagnostic. We classify an effect as reliable at
$r_\Delta \ge 0.6$, attenuated at $0.3 \le r_\Delta < 0.6$, and unresolvable below
$0.3$; this classification relies on $r_\Delta$ directly rather than the geometric mean.

\paragraph{Check 1: does alignment exceed background geometry?} We compare
$\cos(\yv_i, \mathbf{v}_i)$ against the cosine computed with a null partner drawn
uniformly from earlier positions in the same sequence. The difference measures
excess alignment beyond shared causal representation geometry.

\paragraph{Check 2: does the training benefit depend on direction?} We train
self-directed and arbitrary-direction arms with identical gate parameterisations
and evaluate each against baseline and against each other across paired random seeds.
The protocol pre-registers the arbitrary-direction comparison as primary.

\paragraph{Contributions.}
We provide the three checks and a reconstruction gate for verifying the
captured head layout (Sections~\ref{sec:method} and \ref{sec:gate}). The
experiments comprise an eight-seed training factorial, geometry measurements
across twelve pretrained models, and reliability and correlation estimates
for three models (Sections~\ref{sec:paired} and
\ref{sec:a2}--\ref{sec:gqa}).

\section{Related Work}

\paragraph{Head importance and pruning.} \citet{michel2019sixteen} and
\citet{voita2019analyzing} show that many attention heads can be pruned with
limited degradation on their evaluated tasks. Their results concern the
importance of whole heads. We instead intervene along one direction within
each head and use an arbitrary-direction arm to test the specificity of that
choice.

\paragraph{Circuit-level accounts.} \citet{elhage2021mathematical} decompose
attention into $QK$ and $OV$ circuits and treat the diagonal of the attention
pattern as one contribution to the output. Projecting the full head output
onto its own value direction can also remove aligned contributions from
earlier tokens. It therefore differs from deleting only the diagonal term.

\paragraph{Representation geometry.} \citet{ethayarajh2019contextual} finds
anisotropy in contextual representations from ELMo, BERT, and GPT-2. This
motivates a geometric control; it does not establish the geometry of every
head examined here. Check~1 measures that background using values from
earlier positions in the same sequence.

\paragraph{Isotropy is not universal.} \citet{timkey2021bark} show that a
small number of high-variance coordinates dominate cosine similarity in
contextual spaces, so a raw cosine can be large without reflecting shared
structure. \citet{machina2024anisotropy} further report that the large Pythia
models have approximately isotropic embedding spaces, which bounds how much a
generic anisotropy argument can explain. Our null is computed within the same
run, at one stated sequence length, on the same documents as the observed
statistic, so it measures the earlier-position background of each model
directly rather than assuming a uniform anisotropy across the ladder. Across
the multi-head ladder, the earlier-position null accounts for $48.1\%$ to
$63.1\%$ of $\cos(\yv_i, \mathbf{v}_i)$. Even at $6.9$B parameters, $58.1\%$
of the alignment ($0.198$ of $0.340$) remains attributable to the null,
showing that background geometry persists at billion-parameter scales rather
than vanishing with model capacity.

\paragraph{Grouped-query attention.} \citet{ainslie2023gqa} share key and value
projections across groups of query heads. We compare own-group and
neighbouring-group values to assess the statistic's dependence on that sharing in
Section~\ref{sec:gqa}.

\section{Method}
\label{sec:method}

\subsection{The intervention}

Let head $h$ at layer $\ell$ produce output $\yv_i \in \mathbb{R}^{d_h}$ at
position $i$, and let $\mathbf{v}_i$ be the value vector that same head writes
at $i$. Write $\vhat_i = \mathbf{v}_i / \lVert \mathbf{v}_i \rVert$. The
self-directed removal, which we call XSA, replaces the head output with
\begin{equation}
\zv_i \;=\; \yv_i \;-\; \tanh(\alpha)\,\langle \yv_i, \vhat_i\rangle\,\vhat_i ,
\label{eq:xsa}
\end{equation}
where $\alpha$ is a gate for each layer and head. The edit acts before the
output projection. For frozen models, we set the multiplier to one and
intervene on one head at a time, excluding the first token. In the trained
factorial, gates start at zero and are learned jointly with the model; all
positions are included. A zero gate disables the edit, and a negative gate
adds an aligned component. The implementation stabilises normalisation by
using $\mathbf{v}_i/(\lVert\mathbf{v}_i\rVert+\epsilon)$, with
$\epsilon=10^{-8}$ for frozen measurements and $10^{-6}$ for training.
Equation~\ref{eq:xsa} states the idealised unit-direction form.

The \textsc{random} control uses the same gated edit with a fixed arbitrary
direction for each layer and head. Directions are sampled by normalising
isotropic Gaussian vectors, seeded by layer index, and held fixed across
positions, training steps, and run seeds. Both arms use the same normalisation
and gate parameterisation, but their projection magnitudes need not match.
The harness also implements \textsc{meanval} and \textsc{diagmask}
(Appendix~\ref{app:arms}); neither appears in the reported three-arm factorial.

\subsection{Check 0: resolvability}

Split the evaluation corpus into halves $A$ and $B$ and measure each head's
effect $\Delta$ and structural statistic on both halves. Spearman
correlations across heads give the split-half reliabilities $r_\Delta$ and
$r_{\text{stat}}$. The archived analysis applies no Spearman--Brown
correction and reports:
\begin{equation}
\rho_{\max} \;=\; \sqrt{r_\Delta \, r_{\text{stat}}},
\qquad
\rho_{\text{disatt}} \;=\; \rho_{\text{raw}} / \rho_{\max}.
\label{eq:ceiling}
\end{equation}
We classify the effect as reliable at $r_\Delta \ge 0.6$, attenuated at
$0.3 \le r_\Delta < 0.6$, and unresolvable below $0.3$. An unresolvable
effect does not support interpreting a small correlation as absence of an
association. Each structural statistic uses its own reliability estimate.

For $\rho_{\text{raw}}$, the statistic and effect are each averaged across
the two halves before computing their Spearman correlation across heads.
Because the denominator in Equation~\ref{eq:ceiling} uses half-sample
reliabilities, the adjusted value is a diagnostic, not a calibrated estimate
of the latent correlation between the pooled measurements. We retain the
archived notation $\rho_{\max}$, but do not interpret it as a strict bound
on those pooled correlations.

\subsection{Check 1: the anisotropy null}

For each head and position, $\cos(\yv_i, \mathbf{v}_i)$ is compared against
$\cos(\yv_i, \mathbf{v}_j)$ with $j$ drawn uniformly from
$\{1,\dots,i-1\}$ using one-based indexing. The first token is excluded
because it has no earlier partner and its head output equals its own value.
The statistic, \emph{excess}, is the mean paired difference between the
cosines. It controls for alignment with earlier values, not all sources of
anisotropy. The model-ladder figure reports point estimates; the harness
supports layer-cluster bootstrap intervals for head-level data.

\begin{samepage}
Two further diagnostics use different nulls. For attention sinks, we compare
attention mass on the first token with the average mass on the next three
tokens, using queries from the ninth token onward. For massive activations,
we compare the mean maximum absolute hidden coordinate, scaled by the hidden
tensor's standard deviation, with the Gaussian reference
$\sqrt{2\log d}$ for width $d$. This reference is a scale approximation,
not an exact finite-width expected maximum.
\end{samepage}

\subsection{Check 2: matched intervention}
\label{sec:check2}

Check~2 is run as a paired factorial over seeds. For each seed, all arms train
from an identical initialisation on an identical token stream, so the paired
difference controls for variation shared within a seed. Define $\Delta$ as
the arm's validation loss minus baseline loss, so negative values indicate
improvement. The protocol-specified primary endpoint is \textsc{random}
versus baseline. The XSA comparison is secondary, with Holm correction
\citep{holm1979simple} restricted to secondary endpoints. XSA is the only
secondary in the completed factorial, so its adjusted and unadjusted
$p$-values coincide. We report two-sided paired $t$-tests and percentile
bootstrap intervals over seed differences. Separate baseline comparisons
do not constitute a direct test of XSA against \textsc{random}.

The analysis approximates the minimum detectable effect at $80\%$ power and
$\alpha = 0.05$ by
\begin{equation}
\text{MDE} \;=\; 2.9\,\sigma_{\text{paired}} \big/ \sqrt{n},
\label{eq:mde}
\end{equation}
using the observed $\sigma_{\text{paired}}$. We call this the realised MDE
to distinguish it from the planning forecast. It describes sensitivity under
the approximation, not achieved power or a hard threshold for detection.

\subsection{The reconstruction gate}
\label{sec:gate}

A head-layout error can preserve tensor shape while changing the intervention.
We rebuild $\yv = A\,\mathrm{expand\_kv}(V)$ from the captured attention
matrix and values, then compare it with the input to the output projection.
The gate rejects a relative Frobenius error above $10^{-2}$ or a non-finite
value. In each reliability run, it checks the first, middle, and last layers
on one document before measuring head effects.

The implementation notes record an initial Pythia-160M error of
$2.0\times10^{-4}$ and errors of $1.39$ to $1.50$ after reversing heads,
rolling the head axis, or omitting a transpose. These observations motivated
the threshold. In the archived reliability runs, the maximum error across
the checked layers is approximately $1.6\times10^{-3}$ for each model,
below the same threshold. A passed gate supports the tensor reconstruction
at those checked layers; it does not validate the intervention's semantic
interpretation.

\section{Experimental Setup}

\paragraph{Frozen-model measurements.} Check~1 covers twelve pretrained
models, including nine multi-head models from $124$M to $6.9$B parameters
and three grouped-query models. These include the GPT-2
family \citep{radford2019language} and the Pythia suite
\citep{biderman2023pythia}. Model revisions are recorded by hash. The
geometry measurements use WikiText-103 test documents and each run's recorded
block length. Check~0 is restricted to GPT-2, Pythia-160M, and Pythia-410M,
using $64$ documents per half at a fixed length of $128$ tokens. Shorter
documents are dropped rather than padded. This length constraint applies to
the reliability runs, not to the full model ladder.

\paragraph{Trained factorial.} Check~2 uses a decoder-only model with eight
layers, eight heads per layer, and hidden width $512$, approximately $50.9$M
parameters. The data pipeline uses FineWeb-Edu's sample-10BT stream with a
fixed disjoint validation holdout. We trained from scratch on eight seeds
(7, 42, 99, 512, 1337, 2024, 8191, 31337) under three
arms (\textsc{baseline}, \textsc{xsa}, \textsc{random}), giving 24 cells.
Each cell sees $399\,900\,672$ tokens, rounded down from the budget to a
whole number of batches. Runs share initialisation, token order, learning-rate
schedule, and validation batches within each seed. All three arms for a seed
complete before the next seed starts.

\paragraph{Budget.} The completed factorial cost \$15.58 against a \$20
ceiling. The ledger records $16.98$ GPU-hours spent in training cells;
billing covers pod uptime, including time outside those cells. An earlier
$5\times10^{7}$-token pilot is retained separately and is not pooled with
the primary runs.

\section{Results}

\begin{table}[tb]
\centering
\small
\begin{tabular}{@{}lrrrl@{}}
\toprule
Model & $r_\Delta$ & $r_{\text{stat}}$ & $\rho_{\max}$ & Verdict \\
\midrule
GPT-2         & 0.795 & 0.997 & 0.890 & reliable \\
Pythia-160M   & 0.419 & 0.994 & 0.645 & attenuated \\
Pythia-410M   & 0.531 & 0.991 & 0.726 & attenuated \\
\bottomrule
\end{tabular}
\caption{Check~0 split-half reliability, $n = 144$ heads for GPT-2 and
Pythia-160M and $n = 384$ for Pythia-410M, at a fixed sequence length of $128$
tokens with $64$ documents per half. The verdict uses $r_\Delta$;
$r_{\text{stat}}$ here refers to self-cosine. $\rho_{\max}$ is the
attenuation factor in Equation~\ref{eq:ceiling}.}
\label{tab:reliability}
\end{table}

\begin{table}[tb]
\centering
\small
\begin{tabular}{@{}llrrr@{}}
\toprule
Model & Statistic & $\rho_{\text{raw}}$ & $\rho_{\max}$ & $\rho_{\text{disatt}}$ \\
\midrule
\multirow{2}{*}{GPT-2}
  & self-cosine & $+0.462$ & 0.890 & $+0.519$ \\
  & excess      & $+0.279$ & 0.891 & $+0.313$ \\
\multirow{2}{*}{Pythia-160M}
  & self-cosine & $+0.149$ & 0.645 & $+0.231$ \\
  & excess      & $+0.189$ & 0.646 & $+0.292$ \\
\multirow{2}{*}{Pythia-410M}
  & self-cosine & $-0.025$ & 0.726 & $-0.034$ \\
  & excess      & $+0.249$ & 0.727 & $+0.342$ \\
\bottomrule
\end{tabular}
\caption{Spearman correlation between the structural statistic and the measured
behavioural effect, raw and adjusted using Equation~\ref{eq:ceiling}.
The raw correlation uses averages over both halves; the adjustment uses
each statistic's own half-sample reliability.}
\label{tab:a2}
\end{table}

\begin{table}[tb]
\centering
\small
\begin{tabular}{@{}lrr@{}}
\toprule
 & \textsc{random} & \textsc{xsa} \\
 & (primary) & (secondary) \\
\midrule
Mean $\Delta$ (nats)     & $+0.001056$ & $-0.002924$ \\
CI low                   & $-0.000298$ & $-0.003967$ \\
CI high                  & $+0.002390$ & $-0.001709$ \\
$t$                      & $+1.43$     & $-4.67$ \\
$p$                      & $0.197$     & $0.0023$ \\
Holm $p$                 & n/a         & $0.0023$ \\
Cohen's $d_z$            & $+0.504$    & $-1.649$ \\
$\sigma_{\text{paired}}$ & $0.002095$  & $0.001773$ \\
Realised MDE             & $0.00215$   & $0.00182$ \\
\bottomrule
\end{tabular}
\caption{Check~2, eight complete paired seeds, $399\,900\,672$ tokens per run.
Both contrasts are against \textsc{baseline}. Positive $\Delta$ means higher
loss. Intervals are $95\%$ percentile bootstrap intervals over paired seed
differences. The realised MDE uses the observed $\sigma_{\text{paired}}$.
Holm correction does not apply to the primary endpoint.}
\label{tab:paired}
\end{table}

\begin{figure}[tb]
\centering
\includegraphics[width=0.98\linewidth]{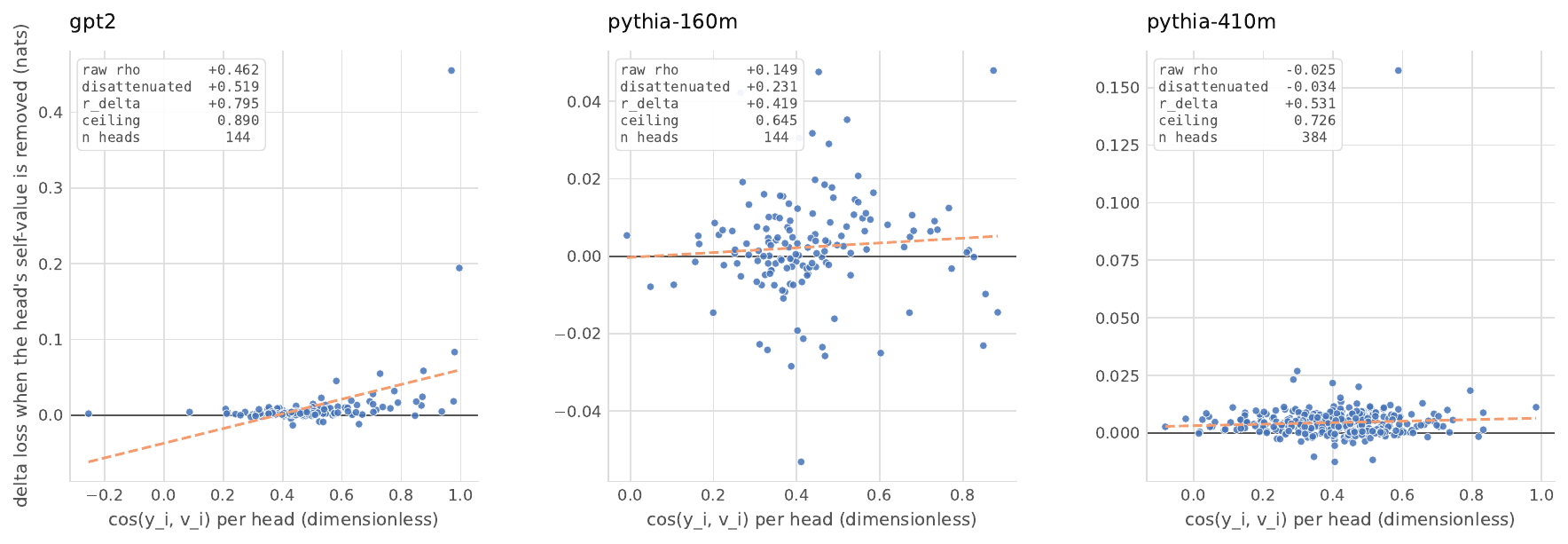}
\caption{Self-cosine against the loss effect of frozen-head intervention.
Points use the half-A statistic and the effect averaged over both halves;
panel annotations report the pooled-statistic correlations in
Table~\ref{tab:a2}. Dashed least-squares lines are visual guides, not tests
of significance. Vertical scales differ across panels.}
\label{fig:a2}
\end{figure}

\begin{figure}[tb]
\centering
\includegraphics[width=0.42\linewidth]{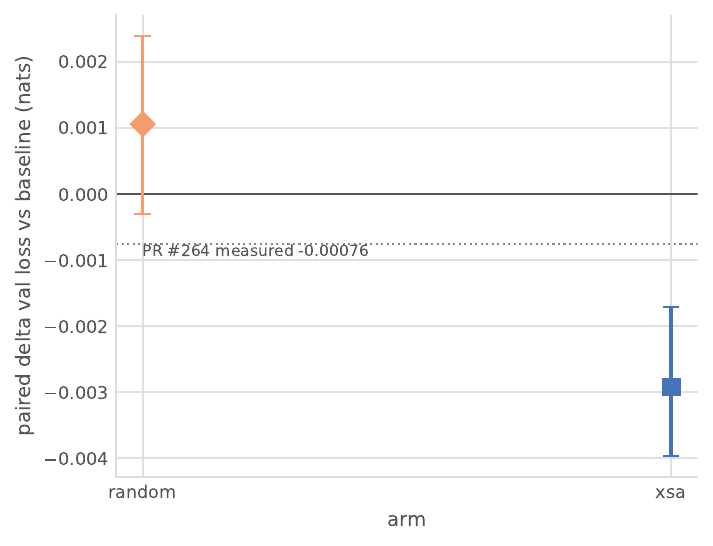}
\caption{Paired difference in validation loss against baseline, with $95\%$
intervals, over eight paired seeds. The \textsc{random} interval crosses zero
and the \textsc{xsa} interval lies below zero. The horizontal reference
line marks an external reference value. The $-0.00076$ value
comes from the H200 comparison in
\href{https://github.com/KellerJordan/modded-nanogpt/pull/264}{PR~\#264},
which uses different training step counts between arms; it is not a matched
replication target for this experiment.}
\label{fig:paired}
\end{figure}

\begin{figure}[tb]
\centering
\includegraphics[width=0.55\linewidth]{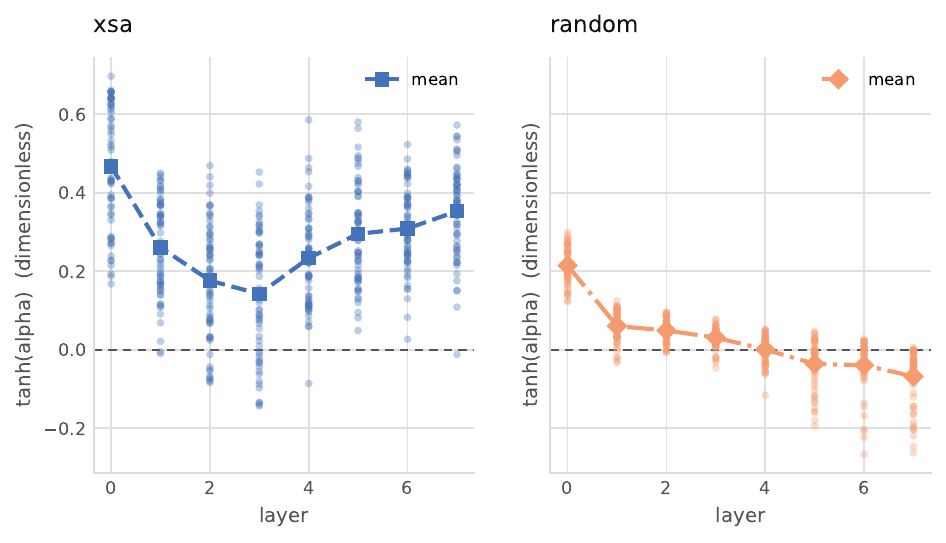}
\caption{Final learned gates $\tanh(\alpha)$ for all heads and seeds, with
markers at the layer mean. XSA has positive means at every layer. The
\textsc{random} mean approaches zero by layer four and becomes negative
in later layers. A negative gate adds the projected component. These are
endpoint measurements, not gate trajectories during training.}
\label{fig:gates}
\end{figure}

\begin{figure}[tb]
\centering
\includegraphics[width=0.50\linewidth]{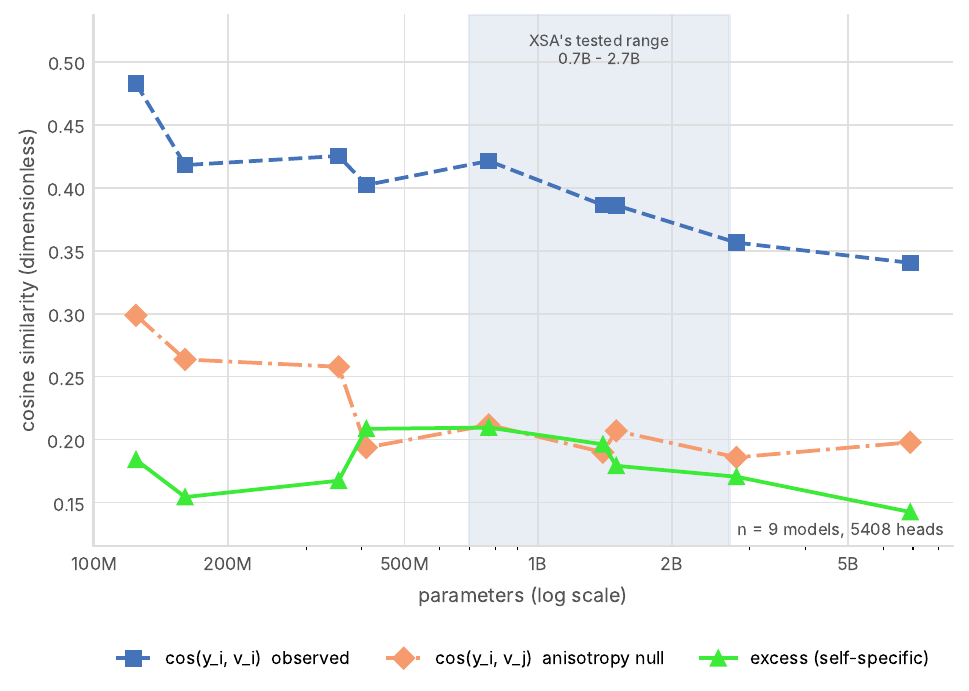}
\caption{Check~1 across the nine multi-head models in the ladder, every model
pinned by revision hash. Observed $\cos(\yv_i,\mathbf{v}_i)$, the anisotropy
null $\cos(\yv_i,\mathbf{v}_j)$, and their difference. The observed curve
generally declines with scale; the excess has no monotone trend.}
\label{fig:ladder}
\end{figure}

\begin{figure}[tb]
\centering
\includegraphics[width=0.48\linewidth]{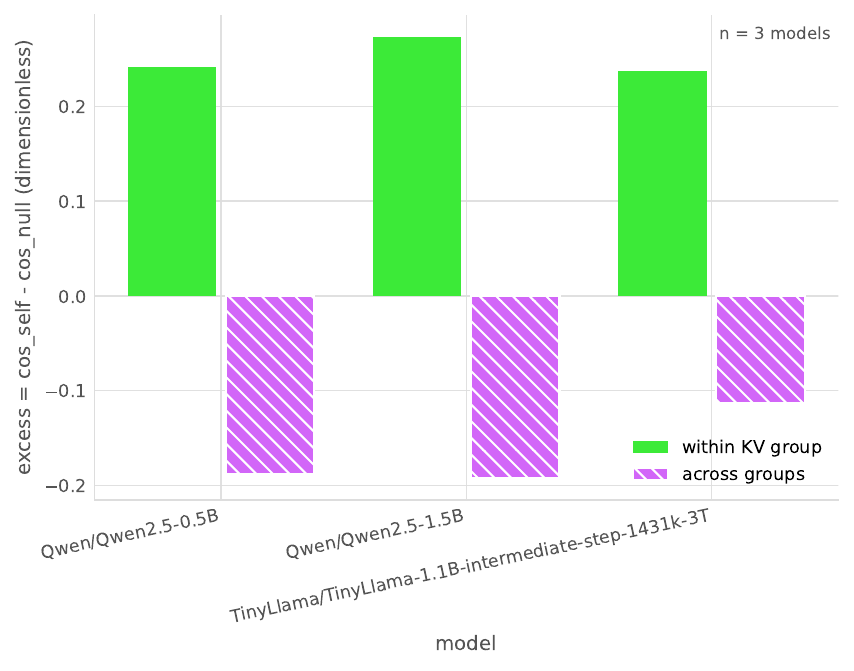}
\caption{Within-group against across-group excess for the three grouped-query
models. Both subtract the own-group earlier-position null. Own-group excess
is positive and neighbouring-group excess negative in all three models.
The harness does not report this grouping contrast for multi-head models.}
\label{fig:gqa}
\end{figure}

\subsection{Resolvability}

Under the Check~0 thresholds, the behavioural effect is reliable in GPT-2
($r_\Delta = 0.795$) and attenuated in Pythia-160M ($0.419$) and Pythia-410M
($0.531$; Table~\ref{tab:reliability}). All three models avoid the unresolvable
category ($r_\Delta < 0.3$). Structural self-cosine reliability is near-perfect
across all three models ($0.991$ to $0.997$). The attenuation is therefore driven
by measurement noise in the single-head intervention effect rather than the
underlying representation statistic. For Pythia-160M, dividing the raw self-cosine
correlation of $0.149$ by the attenuation diagnostic $0.645$ yields $0.231$;
this diagnostic adjustment does not substitute for an adequately resolved behavioural measurement.

\subsection{Structural statistic versus behavioural effect}
\label{sec:a2}

The relation between self-cosine and the measured effect varies by model
(Figure~\ref{fig:a2}, Table~\ref{tab:a2}). GPT-2 has a raw correlation of
$+0.462$ and an adjusted value of $+0.519$; Pythia-410M has an adjusted
value of $-0.034$, close to zero. For excess, the adjusted correlations are
$+0.313$, $+0.292$, and $+0.342$ across the three models. These are
descriptive rank associations, not estimates of explained loss variance or
out-of-sample predictive accuracy.

An earlier Pythia-160M run reported a raw excess correlation of $+0.487$,
compared with $+0.189$ in the current run. Several protocol changes
accompanied the rerun (Appendix~\ref{app:deviations}), so their individual
contributions to this difference are not isolated.

\subsection{The paired factorial}
\label{sec:paired}

All 24 training cells completed, giving eight paired seeds for each contrast
in Table~\ref{tab:paired} and Figure~\ref{fig:paired}.

For the primary endpoint, the \textsc{random} arm's loss difference is
$+0.001056$ nats ($p = 0.197$). Its $95\%$ interval,
$[-0.000298, +0.002390]$, spans both directions, indicating no detectable
change from baseline.

The secondary endpoint, \textsc{xsa} versus baseline, is $-0.002924$ nats with
interval $[-0.003967, -0.001709]$, $t = -4.67$, and Holm-corrected $p =
0.0023$. XSA therefore lowers loss in this training setting. The realised MDE from
Equation~\ref{eq:mde} is $0.00182$ nats for the XSA contrast and $0.00215$ for
the control contrast. The observed XSA effect is about $1.6\times$ its
approximate MDE, whereas the control effect is smaller than its MDE.
Figure~\ref{fig:power} shows sensitivity estimates for the control at both
token budgets.

The benefit concerns a model trained with learned removal. It does not show
that the same component can be removed from an already trained model without
cost. Beyond the separate contrasts against baseline, a direct paired comparison
between \textsc{xsa} and \textsc{random} across the eight seeds confirms that
self-directed removal outperforms the fixed arbitrary control directly
(mean paired difference $-0.003980$ nats, $t = -4.86$, $p = 0.0018$).

Final gates differ between arms (Figure~\ref{fig:gates}). XSA's layer means
range from $0.14$ to $0.47$, indicating partial removal on average.
The control's later-layer means are negative, indicating addition along
its direction. These endpoint measurements do not explain the loss difference.

\subsection{Check 1 across the model ladder}
\label{sec:ladder}

Across nine multi-head models and $5408$ heads, self-cosine generally
declines with model size (Figure~\ref{fig:ladder}). It is $0.483$ for
GPT-2 at $124$M parameters and $0.340$ for Pythia-6.9B. The earlier-position
null also declines overall. Excess remains between $0.14$ and $0.21$
without a monotone trend. These observations do not establish scale
invariance, but show that the raw cosine's trend is partly shared by its
null. The grouped-query models are examined separately below.

Both additional GPT-2 diagnostics exceed their respective nulls
(Figure~\ref{fig:generality}). Attention mass on the first token is
$0.403$, compared with $0.003$ on the matched early positions. The
massive-activation statistic is $12.87$ against a Gaussian reference of
$3.65$, leaving approximately $71.7\%$ after subtraction. Neither pattern
is explained away by its chosen null. The two unexecuted diagnostics are
shown as blank categories.

\subsection{Grouped-query attention}
\label{sec:gqa}

For each grouped-query model, we compare the head output's cosine with its
own group's value and with a neighbouring group's value at the same position.
Both excess values subtract the own-group earlier-position null. For
Qwen2.5-0.5B ($14$ query heads over $2$ key--value heads), own-group excess
is $+0.2415$ and neighbouring-group excess is $-0.1876$. Qwen2.5-1.5B
and TinyLlama-1.1B show the same signs (Figure~\ref{fig:gqa}). The
alignment thus depends on which group's value is used. A negative
across-group excess means alignment below the own-group null, not
necessarily a negative raw cosine. The harness leaves this contrast
undefined for multi-head models, which lack groups sharing a value
projection.

\section{Limitations}

\paragraph{Training and scale.} Check~2 evaluates learned intervention in a
50.9M parameter model trained for $4.0\times10^{8}$ tokens per run. It does
not determine effect signs or magnitudes at larger scales, such as the
0.7B--2.7B regimes evaluated in prior work. Furthermore, the frozen and trained
settings differ in gate constraints and first-token handling, so their outcomes
reflect related but non-identical interventions.

\paragraph{Statistic families.} Both evaluated structural statistics rely on
cosine similarities. Alternative formulations based on attention probabilities
or projection norms may exhibit different empirical associations with head
interventions.

\paragraph{Reliability coverage and correlation scope.} Intervention reliability
was evaluated on three models in the main study, and extended to Pythia-1.4B ($r_\Delta = 0.67$)
and Pythia-2.8B ($0.63$; Appendix~\ref{app:extended_reliability}), where the effect is reliable
under a half-budget protocol ($32$ documents per half). The remaining seven ladder models lack intervention measurements.
Disattenuation applies half-sample reliabilities to pooled correlations, meaning adjusted values
serve as diagnostic bounds alongside raw values rather than corrected ground truth. Rank correlations
describe within-sample monotonicity rather than out-of-sample prediction.

\paragraph{Control scope.} While \textsc{random} and \textsc{xsa} share a
rank-one parametric form, their projected norms and converged gate strengths
diverge. The trained comparison does not isolate projection direction from
magnitude. The control direction is also fixed across positions, whereas
\textsc{xsa} removes a different, input-dependent direction at every token,
so the contrast confounds own-value specificity with per-token versus static
projection. A per-token control, such as projecting out an earlier token's
value as in the Check~1 null, would separate the two. Additionally, reconstruction gates verify captured tensor layouts
and indexing, but do not prove that a value-aligned direction carries solely
self-position information.

\paragraph{Reference mismatch.} The GPT-2 geometry measurement did not
reproduce the project's reference triple within its stated tolerance of
$0.01$. At block $512$ the measured self-cosine, null, and excess were
$0.4828$, $0.2987$, and $0.1840$, against reference values $0.5406$,
$0.3798$, and $0.1608$.

A length sweep accounts for most of that gap. The reference triple is quoted
without a sequence length, and the null is strongly length-dependent while the
observed self-cosine is close to flat. Across $64$ to $1024$ tokens the null
moves by $0.096$, nine times the tolerance, and at $64$ tokens it sits $0.005$
from the reference value, inside it. The excess moves by $0.080$ and comes
within $0.011$ at $256$ tokens. The self-cosine moves by only $0.016$ and
never comes within $0.051$, so length does not account for that component; its
residual gap is consistent with a difference in corpus, truncation, or
head-level aggregation, and we have not identified which.

The consequence for Check~1 is narrow. The null is measured inside the same
runs, at one stated length, on the same documents as the observed statistic,
so the comparison the check makes is internally matched whatever the absolute
level. The excess is positive at every length tested. What the sweep does show
is that a self-specific fraction quoted without a length is not comparable
across studies: the same GPT-2 gives $0.235$ at $64$ tokens and $0.412$ at
$1024$. We therefore report the length everywhere and do not claim
reproduction of the reference triple.

\section{Conclusion}

Learned self-directed removal lowers validation loss by $0.0029$ nats in our
eight-seed factorial ($p = 0.0023$) and outperforms a fixed arbitrary rank-one
control directly by $0.0040$ nats ($p = 0.0018$). The arbitrary-direction control
does not differ from baseline ($+0.0011$ nats, $p = 0.197$). At 50.9M scale the training benefit therefore
requires the own-value direction rather than a fixed arbitrary one; a
per-token control remains to be tested.

However, a within-sequence null shows that background representation geometry
accounts for $48\%$ to $63\%$ of raw self-value alignment across nine models
through 6.9B parameters, and grouped-query attention reverses the alignment
across key--value groups. Furthermore, the intervention effect is reliably
resolvable in GPT-2 but attenuated in Pythia-160M and Pythia-410M, becoming
reliable in Pythia-1.4B and 2.8B under a half-budget protocol. Large internal alignments alone
do not demonstrate representational redundancy. Modifying representations based
on internal statistics requires establishing measurement resolvability,
empirical null comparisons, and matched directional controls.

\subsection*{AI use statement}

AI assistants were used during the development of this work. The authors
defined the research question, experimental design, methodological choices,
and interpretation of results. AI assistants were also used to review the
analysis for errors. The authors take responsibility for every claim,
citation, and number in this paper. Use by category:

\begin{itemize}
\setlength{\itemsep}{2pt}
\item \textbf{Synthetic data.} None. All text comes from public corpora; no
generated data enters any result.
\item \textbf{Hypotheses.} Author-defined. The research question, the three
checks, and the pre-registered primary endpoint were specified by the authors
before any assistant was involved in the experimental programme.
\item \textbf{Theory, proofs, and conceptual frameworks.} None. The paper
contains no theorem or proof, and the conceptual framing of the three checks
is author-originated.
\item \textbf{Methodology and experiment design.} Author-defined. Assistants
enumerated candidate controls and failure modes and drafted the power and
token-budget arithmetic, which the authors reviewed before adoption.
\item \textbf{Code, data analysis, and figures.} Written by the authors. AI
assistants were used for debugging. Every table cell and manifest claim is
recomputed from the stored result files by a checker rather than transcribed,
and every figure is generated from those files; no figure content was drawn
or edited by hand.
\item \textbf{Interpreting results.} Author-owned. Assistants were used
adversarially against the draft, and a second model from a different vendor
reviewed the analysis; both surfaced errors the authors then corrected,
including a reversed sign convention in an earlier draft.
\item \textbf{Writing and editing.} Written by the authors. AI assistants were
used for grammar and language checking.
\item \textbf{Translation.} None. The work was written in English throughout.
\item \textbf{Literature search, references, and structure.} Assistants were
used for literature search and summarisation, reference formatting, and
suggestions on section structure and title wording. The authors verified every
cited work against its original source.
\end{itemize}

\subsection*{Ethics statement}

This work analyses the internal structure of publicly released pretrained
language models and trains small models from scratch on public text. It
introduces no new data collection, no human subjects, and no deployed system.
The small-model training result does not establish that attention components
can be pruned safely in deployed models. Any such use requires evaluation in
the intended model and setting.

\subsection*{Reproducibility statement}

All code, configuration records, raw per-head logs, result files, and
validation scripts are available at \url{https://github.com/Basavarajearlybird/SANITY-CHECKS-FOR-ATTENTION-SURGERY_Original} under the MIT license, and
archived on Zenodo at \url{https://doi.org/10.5281/zenodo.24096410}.
The repository contains per-run factorial records, paired-test summaries,
per-head geometry and intervention measurements, and revision hashes for all
twelve pretrained models. The pilot runs are stored apart from the primary
factorial and are labelled as such.

Every figure is generated from these stored results. The tables are written by
hand, so a checker recomputes each of their 48 cells from the archived result
files and compares against the printed value at the precision shown. Cells
that the manuscript derives are recomputed rather than read back:
$\rho_{\max}$ from its two reliabilities, $\rho_{\text{disatt}}$ from the raw
correlation and that factor, and the realised MDE from the observed paired
standard deviation and seed count. Tables 1--3 currently agree with their
sources on all 48 cells. The checker also reports any table row it does not
know how to verify, so an unchecked row cannot pass silently.

Separately, a manifest recomputes 33 numerical claims from the committed
artifacts and fails on any that does not reproduce within a stated tolerance.
All 33 reproduce from the archived files.

\bibliographystyle{iclr2027_conference}
\bibliography{references}

\clearpage
\appendix

\begin{figure}[tb]
\centering
\includegraphics[width=0.50\linewidth]{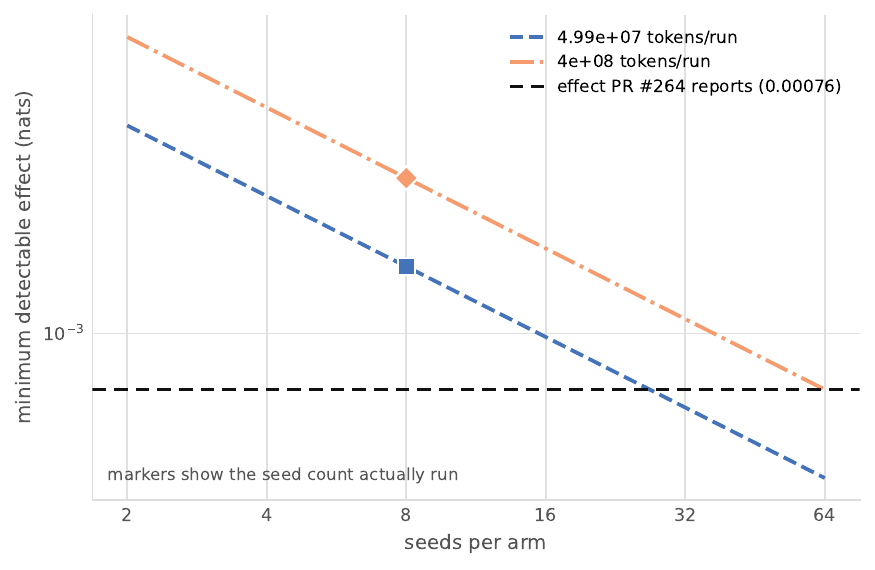}
\caption{Approximate MDE for the \textsc{random}-versus-baseline contrast,
using the observed paired standard deviation at each token budget. Markers
show the eight seeds actually run; the remaining points extrapolate at fixed
variance. Both eight-seed estimates exceed the external reference magnitude
of $0.00076$ nats from the PR~\#264 comparison described in
Figure~\ref{fig:paired}.}
\label{fig:power}
\end{figure}

\begin{figure}[tb]
\centering
\includegraphics[width=0.48\linewidth]{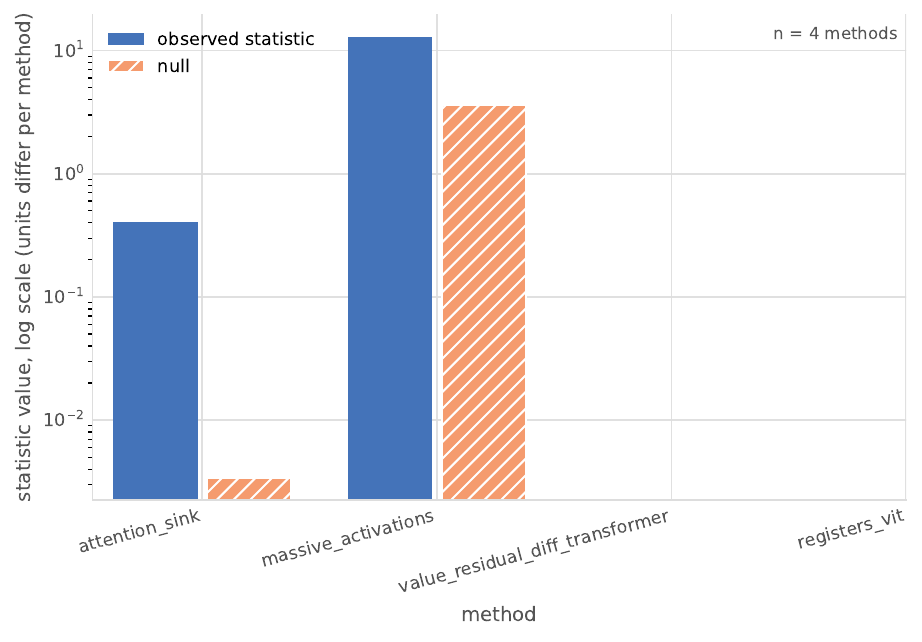}
\caption{Observed statistics and their respective nulls for two GPT-2
diagnostics: attention mass for attention sinks and standard-deviation units
for massive activations. Both exceed their nulls. The logarithmic axis
accommodates different magnitudes; values across diagnostics have different
units. The two blank categories were not run: one requires training matched
models, and the other requires a vision-model harness.}
\label{fig:generality}
\end{figure}

\section{Arm Definitions}
\label{app:arms}

\textsc{baseline} leaves the head output unchanged. \textsc{xsa} uses
Equation~\ref{eq:xsa}, with a learned gate in the factorial and a unit
multiplier in the frozen-head measurements. \textsc{random} substitutes a
fixed arbitrary direction per layer and head and learns the same type of gate.

The two additional harness arms were not evaluated in the primary factorial.
\textsc{meanval} uses a detached exponential moving average of the head's
batch-mean value, updated only during training. \textsc{diagmask} operates
on logits. Its default gated form subtracts $30\tanh(\alpha)$ from diagonal
logits before softmax; the optional hard-mask form sets them to $-\infty$.
Both leave the first token unmasked because it has no alternative key.
These implementations should not be read as completed experimental arms.

\section{Budget Ledger}
\label{app:budget}

The ledger records completed seed blocks and separates training-cell time
from billed pod uptime. Adding cell hours to uptime would count training
twice. An earlier forecast of \$18.16 made that error; the corrected
projection at the same stage was \$13.41. The final billed total was
\$15.58.

\section{Changes to the Measurement Protocol}
\label{app:deviations}

The reliability runs now use a fixed length of $128$ tokens. Earlier runs
allowed the common length to depend on the selected documents: $142$ tokens
at $24$ documents per half and $137$ at $64$. This changes the null partner
set as well as the measurement budget. The analysis also now pools both
halves of the statistic and effect and uses statistic-specific reliability
for disattenuation. The earlier Pythia-160M raw excess correlation of
$+0.487$ and the current value of $+0.189$ belong to different runs;
their difference does not isolate the contribution of any single correction.
Earlier outputs are retained for comparison.

Three departures from the pre-specified plan affect Check~2. The plan
recorded after the pilot called for $12$ seeds per arm; the reported
factorial uses $8$. The planned scale check at a larger training
configuration was not run, so no training result beyond 50.9M parameters is
reported. The \textsc{meanval} and \textsc{diagmask} arms were dropped
before the factorial, leaving \textsc{xsa} as the only secondary endpoint.
The primary endpoint, the token budget band, and the analysis were not
changed.

\section{Extended Reliability at Scale}
\label{app:extended_reliability}

To determine whether the attenuation observed in Pythia-160M ($r_\Delta = 0.419$) and Pythia-410M ($0.531$) is an artifact of smaller parameter counts or persists at scale, we extended Check~0 split-half reliability and Check~0--A2 correlations to \texttt{EleutherAI/pythia-1.4b} ($24$ layers, $16$ query heads, $n = 384$) and \texttt{EleutherAI/pythia-2.8b} ($32$ layers, $32$ query heads, $n = 1024$). Both models were evaluated with the Check~0 protocol at half the document budget ($32$ rather than $64$ documents per half, fixed sequence length of $128$ tokens, evaluated in \texttt{bfloat16} on an NVIDIA RTX 6000 Ada Generation GPU).

Both larger models pass the $r_\Delta \ge 0.60$ threshold and are classified as \textsc{reliable} (Table~\ref{tab:extended_reliability}). Across the Pythia models tested, $r_\Delta$ is $0.419$, $0.531$, $0.666$, and $0.632$; the two larger models are more reliable despite the smaller budget, but the sequence is not monotone and the budgets differ, so we do not claim a scale trend.

\begin{table}[h]
\centering
\small
\begin{tabular}{@{}llrrrrrl@{}}
\toprule
Model & Statistic & $r_\Delta$ & $r_{\text{stat}}$ & $\rho_{\max}$ & $\rho_{\text{raw}}$ & $\rho_{\text{disatt}}$ & Verdict \\
\midrule
\multirow{2}{*}{Pythia-1.4B}
  & self-cosine & \multirow{2}{*}{0.666} & 0.982 & 0.809 & $+0.187$ & $+0.232$ & \multirow{2}{*}{reliable} \\
  & excess      &                        & 0.988 & 0.811 & $+0.249$ & $+0.307$ & \\
\midrule
\multirow{2}{*}{Pythia-2.8B}
  & self-cosine & \multirow{2}{*}{0.632} & 0.967 & 0.782 & $+0.003$ & $+0.003$ & \multirow{2}{*}{reliable} \\
  & excess      &                        & 0.982 & 0.788 & $-0.124$ & $-0.158$ & \\
\bottomrule
\end{tabular}
\caption{Check~0 split-half reliability and Check~0--A2 rank correlations for extended Pythia models ($32$ documents per half, $128$ tokens). $A@V$ reconstruction Frobenius error was $1.66\times 10^{-3}$ on Pythia-1.4B and $1.66\times 10^{-3}$ on Pythia-2.8B. As in Table~\ref{tab:a2}, disattenuation uses each statistic's own split-half reliability.}
\label{tab:extended_reliability}
\end{table}

The association between representation statistics and intervention outcomes varies across models. In Pythia-1.4B, the null-corrected excess cosine ($\rho_{\text{disatt}} = +0.307$) outperforms raw self-cosine ($+0.232$) in predicting intervention effect, mirroring Pythia-410M. In Pythia-2.8B, raw self-cosine completely decouples from intervention effect ($\rho_{\text{raw}} = +0.003, \rho_{\text{disatt}} = +0.003$), despite high measurement reliability ($r_\Delta = 0.632$), and the excess correlation is negative ($\rho_{\text{disatt}} = -0.158$), so the advantage of excess over self-cosine seen in Pythia-410M and 1.4B does not hold here. This indicates that high representation-level cosine alignment does not reliably index intervention sensitivity.

\end{document}